%% file: main.tex
\documentclass{article}
\usepackage{spconf,amsmath,graphicx}
\usepackage{times}
\usepackage{latexsym}
\usepackage[T1]{fontenc}
\usepackage[utf8]{inputenc}
\usepackage{microtype}
\usepackage{CJKutf8}
\usepackage{booktabs}
\usepackage{multirow, makecell}
\usepackage{graphicx}
\usepackage{caption}
\usepackage{subcaption}
\usepackage{bm}
\usepackage{amsmath}
\usepackage{url}

\title{AISHELL-NER: Named Entity Recognition from Chinese Speech}
\name{Boli Chen, Guangwei Xu, Xiaobin Wang, Pengjun Xie, Meishan Zhang, Fei Huang}
\address{
  DAMO Academy, Alibaba Group, China \\
  \{boli.cbl, kunka.xgw, xuanjie.wxb, chengchen.xpj, f.huang\}@alibaba-inc.com \\
  mason.zms@gmail.com
}

\begin{document}

\maketitle

\begin{abstract}
  Named Entity Recognition (NER) from speech is among Spoken Language Understanding (SLU) tasks, aiming to extract semantic information from the speech signal. NER from speech is usually made through a two-step pipeline that consists of (1) processing the audio using an Automatic Speech Recognition (ASR) system and (2) applying an NER tagger to the ASR outputs. Recent works have shown the capability of the End-to-End (E2E) approach for NER from English and French speech, which is essentially entity-aware ASR. However, due to the many homophones and polyphones that exist in Chinese, NER from Chinese speech is effectively a more challenging task. In this paper, we introduce a new dataset AISEHLL-NER for NER from Chinese speech. Extensive experiments are conducted to explore the performance of several state-of-the-art methods. The results demonstrate that the performance could be improved by combining entity-aware ASR and pretrained NER tagger, which can be easily applied to the modern SLU pipeline. The dataset is publicly available at \url{github.com/Alibaba-NLP/AISHELL-NER}.
\end{abstract}


\input{01_introduction}
\input{02_methods}
\input{03_the_new_dataset}

\input{04_experiments}
\input{05_conclusion}

\clearpage
{
  \section{References}
  \ninept
  \bibliographystyle{IEEEbib}
  \bibliography{custom}
}

\end{document}

%% file: 01_introduction.tex
\section{Introduction}

As a fundamental NLP task, Named Entity Recognition (NER) aims to extract Named Entity (NE) into predefined categories, such as \textit{person} (PER), \textit{location} (LOC), \textit{organization} (ORG), \textit{etc}. NER has been mostly studied in the context of written language and traditionally solved as a text-based sequence labeling task, where an NER tagger jointly predicts the NE boundary and category. Recently, NER from speech has been actively studied among Spoken Language Understanding (SLU)~\cite{caubriere-etal-2020-named}. It aims to extract semantic information from speech and has many applications. For instance, muting sensitive NEs such as patient names in audio medical recordings~\cite{cohn-etal-2019-audio}.

\begin{figure}[t]
    \centering
    \begin{subfigure}[b]{.98\linewidth}
        \centering
        \includegraphics[width=\linewidth]{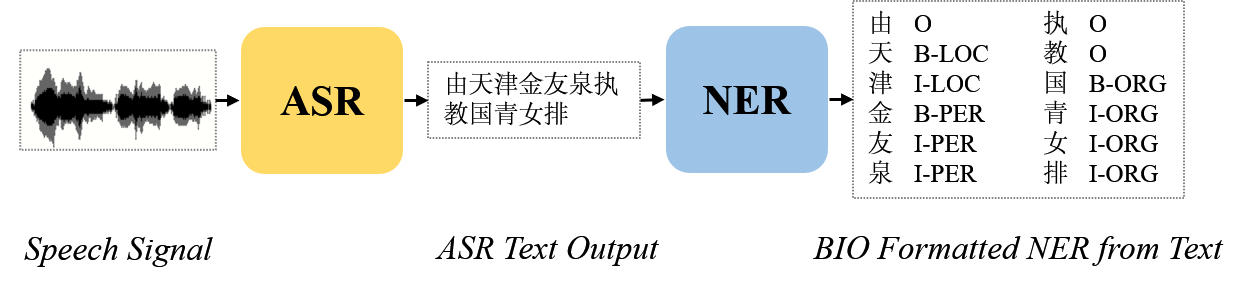}
        \caption{Pipeline}
        \label{fig:pipeline}
    \end{subfigure}
    \par\bigskip 
    \begin{subfigure}[b]{.7\linewidth}
        \centering
        \includegraphics[width=\linewidth]{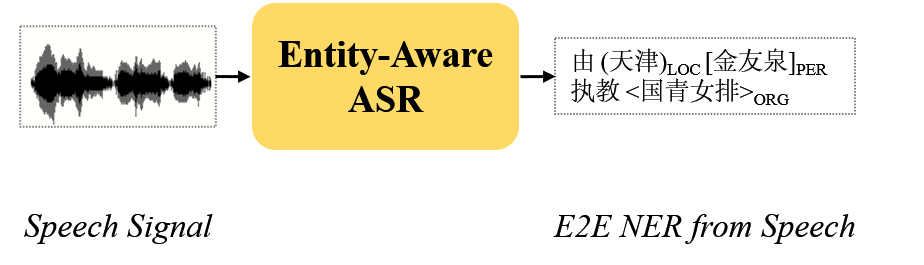}
        \caption{E2E}
        \label{fig:e2e}
    \end{subfigure}
    \caption{(A) Pipeline and (B) E2E methods for NER from Chinese speech.}
    \label{fig:ner_from_speech_methods}
\end{figure}

NER from speech is usually made through a two-step pipeline approach~\cite{hatmi2013named}. As shown in Figure~\ref{fig:pipeline}, an Automatic Speech Recognition (ASR) system first processes the speech signal and outputs the corresponding text in the pipeline, then an NER tagger predicts on the text produced previously by the ASR system. Therefore, the prediction of the NER tagger will be largely influenced by the quality of the speech transcription. Whereas the output text of the ASR system might contain errors or noises, and the imperfect transcription of speech could have a major impact on the final results~\cite{jannet2015evaluate}.

Lately, the End-to-End (E2E) method for NER from French~\cite{8639513} and English~\cite{Yadav2020} speech has been proposed. They effectively adopt entity-aware ASR to predict NEs during the decoding process. As shown in Figure~\ref{fig:e2e}, Special tokens are added to the ASR vocabulary list to annotate NEs in the transcriptions, \textit{i.e.}, $[\ ]$ for PER, $(\ )$ for LOC, and $<\ >$ for ORG. And the entity-aware ASR learns an alignment between the speech signal and its annotated transcription. It could directly extract NEs from speech and eliminate cascading errors in the pipeline. Results on the French and English datasets show that the E2E method could obtain better performance than the pipeline conditioned on the LSTM-based models, which are below current state-of-the-arts. In this paper, we explore the state-of-the-art ASR models based on Transformer~\cite{8682586} and Conformer~\cite{Gulati2020}, as well as NER in the pipeline based on pretrained BERT~\cite{devlin-etal-2019-bert}. Additionally, the French and English datasets use a combination of several corpora, files that do not have any NEs are also removed in \cite{Yadav2020}. AISEHLL-NER is built on the widely used Chinese ASR benchmark, thus it is more suitable for evaluating the SLU capability of ASR systems.

NER from Chinese speech is more challenging, since word segmentation could affect the results~\cite{10.3115/1119384.1119393}. Moreover, there are a lot of homophones and polyphones in the Chinese characters. For example,
\begin{CJK*}{UTF8}{gbsn}
    ``津''\ (Jin)
\end{CJK*}
might refer to the city \textit{Tianjin} (LOC) for short, or just \textit{saliva} (not NE) according to the context. Besides,
\begin{CJK*}{UTF8}{gbsn}
    ``金''\ (Jin)
\end{CJK*}
is pronounced exactly the same as
\begin{CJK*}{UTF8}{gbsn}
    ``津''\ (Jin),
\end{CJK*}
while it usually means \textit{golden} (not NE), and sometimes can be a \textit{ person name} (PER). Consequently, the correct transcription is especially important for NER from Chinese speech.

Many existing works focus on Chinese NER from textual data in the literature~\cite{zhang-yang-2018-chinese, li-etal-2020-flat}, while NER from Chinese speech has not been widely studied due to the lack of a publicly available dataset. To this end, we present a new dataset for this topic, namely AISHELL-NER. It is built upon the open-source Chinese speech corpus AISHELL-1~\cite{8384449}. The annotation process is detailed in Section~\ref{sec:4}.

We also propose to combine entity-aware ASR and pretrained NER tagger to improve the performance for NER from speech. Extensive experiments conducted on AISHELL-NER demonstrate the superiority of our proposed method. Since the modern SLU system in industrial applications usually adopts the pipeline approach~\cite{Ruan2020}, our proposed method could also be easily applied to this configuration.

Our contributions are summarized as follows:

\begin{itemize}
    \item We present the publicly available dataset for NER from Chinese speech AISHELL-NER, with the annotations for three NE categories: PER, LOC, and ORG.
    \item AISHELL-NER is more suitable for evaluating the SLU capability of ASR systems than the existing spoken NER corpora. It can also be used as a text-based NER task.
    \item Extensive experiments are conducted to compare several state-of-the-art ASR and NER models. The results demonstrate the importance of pretraining for NER.
    \item We propose to improve the performance by combining entity-aware ASR and pretrained NER tagger, which can be easily applied to the modern SLU pipeline.
\end{itemize}

%% file: 02_methods.tex
\section{Methods}

\subsection{ASR}

Since Transformer has been shown to be more effective than RNN/LSTM for ASR on various corpora~\cite{9003750}, we build the ASR components with Transformer-based encoder-decoder architecture. The encoder maps the input sequence of features $\bm{x} = (x_1, ..., x_n)$ to a sequence of hidden states $\bm{h} = (h_1, ..., h_n)$. The Transformer encoder is composed of $N$ stacked identical layers, where each layer consists of two sub-layers, \textit{i.e.}, multi-head attention and position-wise feed-forward network. Residual connections are employed around the sub-layers, followed by a layer normalization. Positional embeddings are added to inject positional information. Additionally, the Transformer encoder could be enhanced with CNN, namely Conformer~\cite{Gulati2020}. The Conformer encoder stacks a convolution subsampling layer before self-attention to process the speech signal. Given the hidden states $\bm{h}$, the decoder generates the output sequence one by one. The structure of the Transformer decoder is similar to the Transformer encoder, except that a third sub-layer is inserted to perform multi-head attention over the output of the encoder. The values corresponding to illegal connections are all masked in the self-attention layers of the decoder, in order to preserve the auto-regressive property and prevent rightward information flow in the decoder. The decoding process could be enhanced with the joint CTC/attention approach~\cite{hori-etal-2017-joint} and the Transformer-based Language Model (LM)~\cite{8462682}. The speech augmentation approaches could also be applied to improved the robustness for ASR, such as \textit{speed perturbation} and \textit{SpecAugment}~\cite{Park2019}.

ASR in the NER from speech pipeline is task-agnostic. The E2E method uses entity-aware ASR, which does not require modifying the ASR framework. It adds special tokens to the ASR vocabulary list to identify NEs, \textit{i.e.}, $[\ ]$ for PER, $(\ )$ for LOC, and $<\ >$ for ORG. The special token pair is assigned before and after an NE to identify it. Entity-aware ASR learns an alignment between the speech and the annotated transcription. During the decoding process, NEs in the transcriptions are bounded by their corresponding special token pair. Thus NEs can be predicted by the entity-aware ASR during decoding in an E2E manner.

\subsection{NER}

NER is usually solved as a sequence labeling task on textual documents. BERT~\cite{devlin-etal-2019-bert} are commonly used as the NER tagger. Pretraining has made it possible to effectively utilize the high capacity of Transformer architectures for NER, leading to current state-of-the-arts. Given a tokenized sequence $\bm{c} = (c_1, ..., c_m)$ of length $m$ as the input, the NER tagger outputs the probability distribution over the NER categories $l(c)$ as well as the final sequence of NER tags $\hat{\bm{l}} = (\hat{l}_1, \dots, \hat{l}_m)$. Softmax~\cite{doi101162} is used by BERT to model the output distribution over the category labels. Since NEs might span several consecutive tokens within a sentence, special tagging schemes are often employed for the labeling process, \textit{e.g.}, BIO, where the \textit{\textbf{B}eginning}, \textit{\textbf{I}nside} and \textit{\textbf{O}utside} sub-tags are distinguished. This tagging scheme could introduce strong dependencies between subsequent tags.

%% file: 03_the_new_dataset.tex
\begin{table}[tbp]
    \centering
    \ninept
    \begin{tabular}{@{}lccccc@{}}
        \toprule
              & \# Sentence & w/ NE  & \# PER & \# LOC & \# ORG \\ \midrule
        Train & 120,098     & 40,839 & 15,842 & 20,693 & 21,455 \\
        Dev   & 14,326      & 5,173  & 1,902  & 2,588  & 2,731  \\
        Test  & 7,176       & 2,376  & 898    & 1,330  & 1,165  \\ \bottomrule
    \end{tabular}
    \caption{Statistics of AISHELL-NER.}
    \label{table:stat}
\end{table}

\section{The New Dataset}
\label{sec:4}

We introduce the dataset AISHELL-NER in this section. It is based on AISHELL-1~\cite{8384449}, an open-source speech corpus that contains over 170 hours of Mandarin speech data and has been commonly used for evaluating the performances of ASR systems on Chinese. The corpus covers five domains: “Finance”, “Science and Technology”, “Sports”, “Entertainments” and “News”. And many NEs can be extracted from the transcriptions. This corpus is published under Apache License v.2.0 by Beijing Shell Shell Technology Co., Ltd, allowing for redistribution with or without modifications. The conditions for redistribution and the participants' privacy rights are well respected during the data collection process.

\subsection{Annotation Process}
\label{sec:4_annotation}

We follow the previous work~\cite{Yadav2020} and annotate three types of NE: PER, LOC, and ORG. We first finetune a pretrained Chinese BERT-base model on MSRA~\cite{levow-2006-third}, which is a commonly used Chinese NER dataset that contains these three categories. The model achieves a 93.95 F1 score on the test set. We then use the finetuned model as the NER tagger to pre-tag the transcriptions of AISHELL-1, providing an appropriate start point for later manual annotation. 70 annotators and 10 experienced experts follow the Chinese annotation guidelines for entities of MSRA to manually re-annotate the pre-tagged transcriptions. Since the articles in AISHELL-NER belong to common domains, all the annotators have adequate linguistic knowledge and educational/cultural background. To eliminate biases during the annotation process, they are instructed with detailed annotation principles. The transcriptions are manually annotated and submitted in batches, and each batch contains no more than 5000 sentences. To ensure the quality of manual annotations, we randomly select 10\% sentences in each batch and process the verification. The experienced experts check for wrong or omissive annotations and make sure the sentence-level accuracy of each batch is higher than 97\%, otherwise, the batch will be re-annotated.

\subsection{Corpus Statistics}
\label{sec:4_stats}

Table~\ref{table:stat} gives the numbers of sentences that contain NE and the detailed statistics of NE numbers for each category in AISHELL-NER. The default train, dev and test split in AISHELL-1 is adopted. There are 48,388 sentences that contain a total number of 68,604 NEs in our annotation, and nearly 2/3 of the sentences do not contain any NE. This might make it hard for the E2E method for NER from speech to recognize NEs. In addition to annotating the transcriptions with special tokens, we also provide a two-column BIO format version similar to MSRA. This allows for training the text-based NER tagger in the pipeline, as well as using AISHELL-NER for a text-based NER task (results discussed in Section~\ref{sec:5_ner}).

%% file: 04_experiments.tex
\section{Experiments}
\label{sec:5}

\subsection{Models}

The ASR models is built with ESPnet~\cite{Watanabe2018}. To investigate the importance of ASR performance for the pipeline, we compare a Transformer-based ASR model with a Conformer-based one. The Transformer ASR only uses the attention-based encoder-decoder architecture. While the Conformer ASR makes use of both the joint CTC/attention decoding and LM. Both Transformer and Conformer Encoder have 4 attention heads with 2048 linear units. They stack 12 layers of encoder blocks, and their output size is 256. The decoders use the same configuration, except that they have only 6 layers. SpecAugment is used to randomly apply time and frequency masking blocks to the log mel-filterbank features. Speed perturbation with factors of 0.9, 1.0 and 1.1 is also applied to Conformer, resulting in 3-fold data augmentation. Label smoothing with the weight set to 0.1 is also used. The CNN module of the Conformer encoder uses a kernel size of 15. They are trained with the Adam optimizer. The learning rate is set to 0.002 for Transformer ASR and 0.0005 for Conformer ASR with a linear warmup scheduler. During decoding, the width of the beam search is set to 5. The modeling units for them are characters in our experiments.

The NER tagger in the pipeline is trained on the BIO formatted text. It adopts the configuration of BERT-base, \textit{i.e.}, 12 layers of 12 attention heads and the hidden size of 768. The BERT-based NER tagger is optimized by AdamW~\cite{loshchilov2018decoupled}, with the learning rate set to 0.00005 and a linear warmup scheduler. Note that the EA-ASR models have difficulty to benefit from text-based pretraining like BERT. For fair comparisons between them, we conduct experiments using both NER taggers with and without pretraining. The NER taggers that use pretraining are denoted by PT.

\subsection{ASR Results}
\label{sec:5_asr}

We first compare the Character Error Rate (CER) of the ASR and Entity-Aware ASR (EA-ASR). EA-ASR is trained on AISHELL-NER with NEs annotated by special tokens. When computing CER for EA-ASR on the dev and test set, the special tokens that indicate NEs are ignored. Table~\ref{table:asr_result} gives CER for the ASR and EA-ASR models on the dev and test set. Conformer ASR outperforms Transformer ASR, whereas both Conformer EA-ASR and Transformer EA-ASR are faced with a small loss in performance, \textit{i.e.}, 0.09\% of CER for Transformer EA-ASR and 0.04\% for Conformer EA-ASR on the test set. Since Conformer has higher modeling capacity than Transformer, the loss of performance is less significant for Conformer EA-ASR. This suggests that annotating NEs by the special tokens could harm the performance of the ASR system and a more robust approach should be developed.

\begin{table}[tbp]
  \centering
  \ninept
  \begin{tabular}{@{}lccc@{}}
    \toprule
    CER (\%)                     & \multicolumn{1}{l}{} & Dev           & Test          \\ \midrule
    \multirow{2}{*}{Transformer} & ASR                  & 8.05          & 9.16          \\
                                 & EA-ASR               & 8.39          & 9.25          \\ \midrule
    \multirow{2}{*}{Conformer}   & ASR                  & \textbf{4.42} & \textbf{4.75} \\
                                 & EA-ASR               & 4.45          & 4.79          \\ \bottomrule
  \end{tabular}
  \caption{CER (\%) of ASR and Entity-Aware ASR (EA-ASR).}
  \label{table:asr_result}
\end{table}

\subsection{NER Results}
\label{sec:5_ner}

AISHELL-NER can also be used as a text-based NER benchmark. We train NER models based on BERT with and without pretraining on the BIO formatted train set. The trained NER models are evaluated on the ground-truth text of the test set. They are also used in the NER from speech pipeline. We compute the commonly used evaluation metrics for the NER task, \textit{i.e.}, \textit{precision}, \textit{recall}, and \textit{F1 score}. In the two-step pipeline, the ASR models output the speech transcriptions, then the NER taggers predict the NEs. For the EA-ASR (E2E) methods, the NEs can be directly identified by the special tokens.

As shown in Table~\ref{table:ner_result}, pretrained BERT (denoted by PT) generally outperforms the models without pretraining by a large margin. Although the performance of EA-ASR on the ASR task is slightly damaged, both Transformer and Conformer EA-ASR (E2E) outperform the pipelines without pretraining. However, the pipelines using BERT-PT can still obtain higher NER scores than EA-ASR (E2E). This suggests that even though the NER taggers in the pipeline will be affected by error propagation, they can still take advantage of pretraining to boost the performance especially when the ASR system has better performance. Since the input of EA-ASR is audio, it is not designed for pretraining on textual data. Therefore, a pretrainable E2E method for NER from speech could be an interesting research direction.

In addition, combining EA-ASR and pre-trained BERT can further improve the performance of the pipeline. The Conformer EA-ASR and BERT-PT pipeline obtains higher scores than the pipelines with Conformer ASR and the E2E methods. ASR in the SLU pipeline is usually task-agnostic and no information corresponding to the downstream tasks like NER is provided. However, such information might be helpful to choose better partial hypotheses for NER during the decoding process. In some cases the special tokens that identify NEs do not correspond to each other for the E2E methods, \textit{e.g.}, an NE bounded by "(" and "]". This allows the NER tagger to refine the transcriptions and improve the results. The modern SLU system in industrial applications usually uses the pipeline~\cite{Ruan2020}. Thus our proposed method could be easily applied to this configuration.

\subsection{Results of Different NE Categories}

\begin{table}[tbp]
  \centering
  \ninept
  \begin{tabular}{@{}lcccc@{}}
    \toprule
                                                 &         & Precision      & Recall         & F1             \\ \midrule
    Ground-Truth Text                            & BERT    & 81.60          & 82.25          & 81.93          \\
                                                 & BERT-PT & \textbf{93.64} & \textbf{92.89} & \textbf{93.26} \\ \midrule
    \multicolumn{2}{l}{Transformer EA-ASR (E2E)} & 65.32   & 63.39          & 64.34                           \\
    \multicolumn{2}{l}{Conformer EA-ASR (E2E)}   & 74.82   & 71.98          & 73.37                           \\ \midrule
    Transformer                                  & BERT    & 65.17          & 63.33          & 64.23          \\
    ASR                                          & BERT-PT & 67.88          & 66.74          & 67.31          \\ \midrule
    Transformer                                  & BERT    & 64.06          & 62.15          & 63.09          \\
    EA-ASR                                       & BERT-PT & 67.12          & 64.83          & 65.95          \\ \midrule
    Conformer                                    & BERT    & 68.81          & 69.48          & 69.14          \\
    ASR                                          & BERT-PT & 74.01          & 74.07          & 74.04          \\ \midrule
    Conformer                                    & BERT    & 69.66          & 69.27          & 69.47          \\
    EA-ASR                                       & BERT-PT & \textbf{75.54} & \textbf{74.27} & \textbf{74.90} \\ \bottomrule
  \end{tabular}
  \caption{NER results for AISHELL-NER.}
  \label{table:ner_result}
\end{table}

\begin{table}[tbp]
  \centering
  \ninept
  \begin{tabular}{@{}lcccc@{}}
    \toprule
                      &     & Precision & Recall & F1    \\ \midrule
    Ground-Truth Text & PER & 95.75     & 95.43  & 95.59 \\
    \ \ \ \ +         & LOC & 94.54     & 95.18  & 94.86 \\
    BERT-PT           & ORG & 90.90     & 88.32  & 89.59 \\ \midrule
    Conformer EA-ASR  & PER & 58.18     & 55.44  & 56.78 \\
    \ \ \ \ +         & LOC & 86.05     & 86.11  & 86.08 \\
    BERT-PT           & ORG & 75.59     & 74.50  & 75.04 \\ \bottomrule
  \end{tabular}
  \caption{Results in different NE categories.}
  \label{table:ners_result_category}
\end{table}

Since there is a gap between NER performances from the speech signal and the ground-truth text. To investigate the possible reasons, we examine the NER scores of different NE categories. Table~\ref{table:ners_result_category} shows that the major loss occurs in PER. In our observation, there are a lot of homophones and polyphones in Chinese, which make it especially hard for recognizing person names from Chinese speech. For example, the EA-ASR model mistakenly recognizes
\begin{CJK*}{UTF8}{gbsn}
  郭京\ (Guo Jing) as 郭晶\ (Guo Jing),
\end{CJK*}
which are pronounced exactly the same in Chinese speech. Whereas it is not a terrible error when the model takes the person name for another one that is pronounced similarly. And the metrics might need to be adapted to consider such situation for a more reasonable evaluation.

%% file: 05_conclusion.tex
\section{Conclusion}

This paper presents the first study about NER from Chinese speech, as well as the publicly available dataset for this task, namely AISHELL-NER. It could directly facilitate the evaluation of the SLU capability. As ASR in the SLU pipeline is usually task-agnostic, we find that the pipeline can be improved by combining entity-aware ASR and pretrained NER tagger. Meanwhile, our proposed method could be easily applied to the SLU pipeline and does not require modification of the model architectures. Since there is still a gap between the performances of NER from the speech and the ground-truth text, the task is worth further research and AISHELL-NER could be a proper benchmark.